\documentclass[conference]{IEEEtran}
\usepackage{times}

\usepackage[numbers]{natbib}
\usepackage{multicol}
\usepackage[bookmarks=true]{hyperref}
\hypersetup{
	colorlinks=true,
	linkcolor=orange,
	filecolor=magenta,      
	urlcolor=orange,
	citecolor=orange,
}
\usepackage{url}
\usepackage{float}
\usepackage{microtype}
\usepackage{graphicx}
\usepackage{subfigure}
\usepackage{booktabs} %
\usepackage{caption} %
\usepackage{wrapfig}

\usepackage{amsmath}
\usepackage{amssymb}
\usepackage{mathtools}
\usepackage{amsthm}

\usepackage[capitalize,noabbrev]{cleveref}
\usepackage[most]{tcolorbox}
\tcbset{
  promptbox/.style={
    enhanced,
    colback=gray!4,
    colframe=black!55,
    boxrule=0.6pt,
    arc=6pt,          %
    left=10pt,right=10pt,top=8pt,bottom=8pt,
    fonttitle=\bfseries,
    coltitle=black,
    attach boxed title to top left={xshift=8pt,yshift=-2mm},
    boxed title style={
      colback=white,
      colframe=black!55,
      arc=4pt,
      boxrule=0.6pt,
      left=6pt,right=6pt,top=2pt,bottom=2pt
    }
  }
}

\newtcolorbox{PromptBox}[1]{promptbox,title={#1}}
\theoremstyle{plain}

\theoremstyle{definition}

\theoremstyle{remark}

\usepackage[textsize=tiny]{todonotes}

\def \Acronym {\textsc{LongPi}\texttrademark}
\def \Acronym {$\pi_{0.6}^\text{mem}$}
\def \Acronym {\texttt{MEM}}

\def \MethodName {$\textit{Multi-Scale Embodied Memory}$}

\def \pizerosix {$\pi_{0.6}$}
\def \nchallengerollouts {10}  %

\IEEEoverridecommandlockouts

\begin{document}
\makeatletter

\let\@oldmaketitle\@maketitle%
\renewcommand{\@maketitle}{\@oldmaketitle%
  \begin{center}
  \captionsetup{type=figure}
  \includegraphics[width=1.0\textwidth]{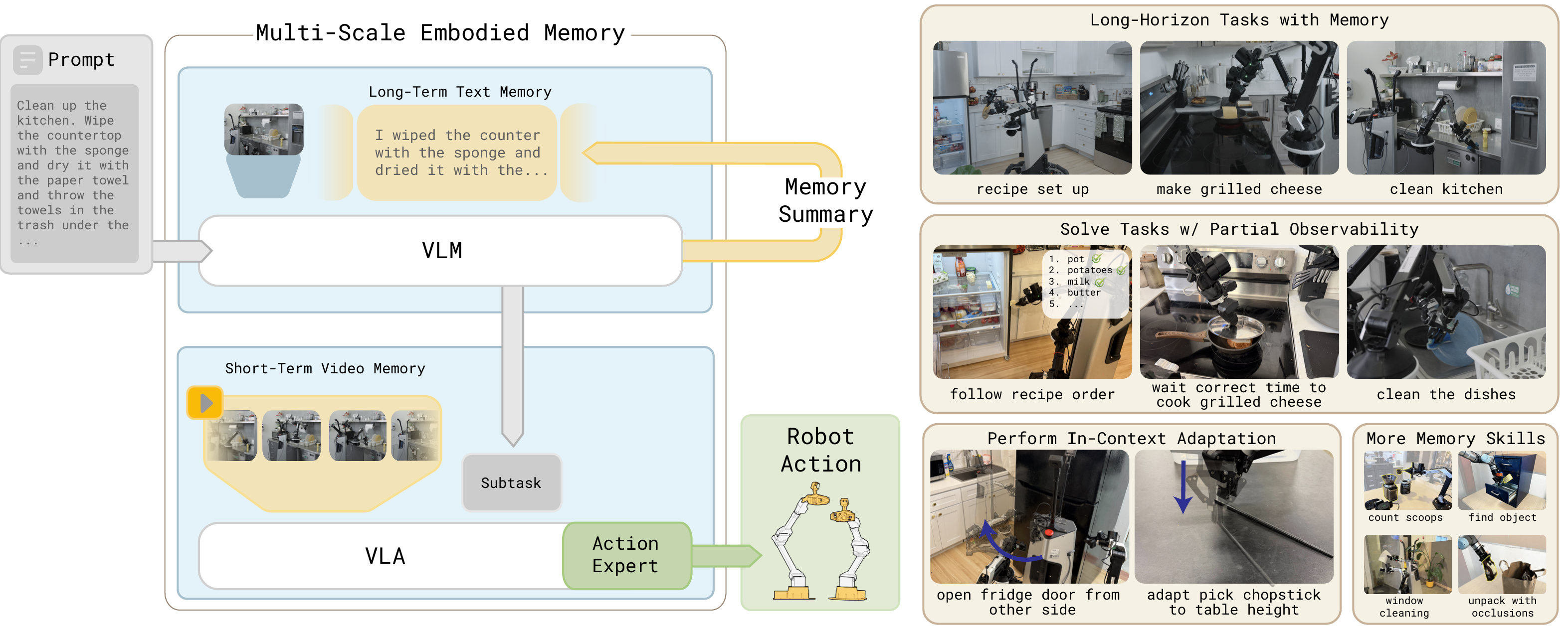}
  \caption{\small{\MethodName{} (\Acronym) equips Vision Language Action Models (VLAs) with memory for solving long-horizon tasks at scale. It has two key components: an efficient video encoder for short-horizon image-based memory, and a language-based memory mechanism for capturing long-horizon memory. After training on a diverse corpus of robot and non-robot data, \Acronym{} VLAs can solve tasks that require up to fifteen minutes of memory, handle partial observability, and perform in-context adaptation of manipulation strategies.}}
    \label{fig:teaser}
    \end{center}
}
\makeatother

\title{\texttt{MEM}: Multi-Scale Embodied Memory for\\Vision Language Action Models}

\author{
  Marcel Torne$^{\ast\,1\,2\,\dagger}$ \quad
  Karl Pertsch$^{\ast\,1}$ \quad
  Homer Walke$^{3\,\dagger}$ \quad
  Kyle Vedder$^{1}$ \quad
  Suraj Nair$^{1}$ \quad
  Brian Ichter$^{1}$ \\
  Allen Z. Ren$^{1}$ \quad
  Haohuan Wang$^{1}$ \quad
  Jiaming Tang$^{1\,4\,\dagger}$ \quad
  Kyle Stachowicz$^{1\,3\,\dagger}$ \quad
  Karan Dhabalia$^{1}$ \quad
  Michael Equi$^{1}$ \\
  Quan Vuong$^{1}$ \quad
  Jost Tobias Springenberg$^{1}$ \quad
  Sergey Levine$^{1}$ \quad
  Chelsea Finn$^{1}$ \quad
  Danny Driess$^{1}$ \\[4pt]
  \url{https://pi.website/research/memory}
  \\[4pt]
}

\maketitle
\addtocounter{figure}{-1} %

\renewcommand{\thefootnote}{}%
\footnotetext{
$^*$equal contribution, $^\dagger$work conducted during an internship at PI.\\
$^{1}$Physical Intelligence, $^{2}$Stanford University, $^{3}$UC Berkeley, $^{4}$MIT
}

\begin{abstract}

Conventionally, memory in end-to-end robotic learning involves inputting a sequence of past observations into the learned policy. However, in complex multi-stage real-world tasks, the robot's memory must represent past events at multiple levels of granularity: from long-term memory that captures abstracted semantic concepts (e.g., a robot cooking dinner should remember which stages of the recipe are already done) to short-term memory that captures recent events and compensates for occlusions (e.g., a robot remembering the object it wants to pick up once its arm occludes it). In this work, our main insight is that an effective memory architecture for long-horizon robotic control should combine multiple modalities to capture these different levels of abstraction. We introduce \MethodName{} (\Acronym{}), an approach for mixed-modal long-horizon memory in robot policies. \Acronym{} combines video-based short-horizon memory, compressed via a video encoder, with text-based long-horizon memory. Together, they enable robot policies to perform tasks that span up to fifteen minutes, like cleaning up a kitchen, or preparing a grilled cheese sandwich. %
Additionally, we find that memory enables \Acronym{} policies to intelligently adapt manipulation strategies \emph{in-context}.

\end{abstract}

\section{Introduction}
\label{sec:intro}

Efficiently and effectively endowing robotic policies with memory requires multiple levels of abstraction. While in principle we could simply encode the entire sequence of past observations into the context of the policy, this becomes intractable for long tasks, necessitating either very short sequences or significant subsampling.
However, in practical settings, the \emph{representation} required for long- and short-term memory is likely to be very different. For example, the robot might need to remember a recent observation to handle occlusions, and it might need to remember that it has already added one of the ingredients when cooking a meal. But these memories are fundamentally different: the former might require storing a few images over a short time period, the latter might require long-term memory but only a few bits of information.

An effective memory architecture for robot policies should use multiple modalities to represent memories at these different levels of abstraction. For short-horizon memory, dense image-based memory is well-suited to resolve occlusions and allows the robot to quickly adapt its manipulation strategy, e.g., by changing the grasp after failing to pick up an object. For long-horizon memory, we often only need to keep track of events at a \emph{semantic} level, such as which ingredient has already been added to a dish. In this case, a language-based representation provides much better compression than raw observations, and allows us to store high-level memories over long time periods.

Based on these observations, we introduce \MethodName{} (\Acronym{}), a system for equipping policies with multi-modal, long-horizon memory. \Acronym{} combines two key ingredients to make long-horizon memory tractable. 
First, we use a video encoder architecture to effectively encode multiple seconds of dense image-based memory into a compact representation. Second, we introduce a language-based memory mechanism in which the policy keeps track of semantic events in a compressed language format.
This memory system can not only accommodate very long horizon tasks, but also enables a variety of new capabilities by leveraging the short-term memory, such as in-context adaptation to correct mistakes, and resilience to partial observability and self-occlusion. 

To evaluate \Acronym{}, we integrate it into the \pizerosix{} model~\citep{pi06_model_card}, a generalist VLA trained on a diverse mixture of robot, vision-language, and video data.
We show that the resulting policy achieves state-of-the-art performance across a wide range of complex manipulation tasks. We also show that \Acronym{} enables our policy to solve long-horizon tasks like cleaning up a whole kitchen or preparing a grilled cheese sandwich, which require keeping track of memories for up to fifteen minutes.

In summary, we introduce a system for multi-scale, long-horizon memory for robot policies. By effectively representing short- and long-horizon memory via video and language representations respectively, \Acronym{} allows robot policies to keep track of memories across tens of minutes, without sacrificing runtime constraints. To implement \Acronym{}, we use an efficient video encoder architecture and a language-based memory system, and demonstrate their effectiveness through state-of-the-art performance across diverse robot tasks. With \Acronym{}, we enable robot policies to perform complex tasks like cleaning up whole kitchens, which can span up to fifteen minutes.

\section{Related Work}
\label{sec:related_work}

\noindent \textbf{Vision language action models with memory.} 
Recent works have demonstrated that learned robot control policies, trained on large amounts of diverse robot experience, can lead to generalizable manipulation, e.g., in unseen environments~\citep{pi05, blackpi0, geminirobotics, lbmtri2025, nvidia2025gr00tn1openfoundation,ye2026worldactionmodelszeroshot}. Vision language action models (VLAs)~\citep{driess2023palm, rt22023arxiv, open_x_embodiment_rt_x_2023, octo_2023, Doshi24-crossformer, kim2024openvla, black2024pi_0, wen2024tinyvlafastdataefficientvisionlanguageaction, zhen20243dvla, belkhale2024minivla, szot2024multimodal, pertsch2025fast, geminirobotics2025, wen2025dexvla, bjorck2025gr00t, wu2026pragmatic, team2026gigabrain} are a popular approach for training such generalist policies, in which a pre-trained vision-language model is finetuned with robot experience. While many of today's state-of-the-art VLAs are trained without memory and act purely based on the current observation of the environment~\citep{kim2024openvla, blackpi0, pi05, geminirobotics, lbmtri2025, nvidia2025gr00tn1openfoundation}, a growing number of works have explored adding memory to policy training since it is a core requirement for solving a wide range of real-world tasks. While early works explore architectures with recurrent memory modules~\citep{rahmatizadeh2018vision, mandlekar2021matters}, more recent works that use transformer-based architectures simply pass a dense history of prior observations into the policy~\citep{shafiullah2022behavior, lee2024behavior, octo_2023}; computational and latency constraints make it challenging to scale such approaches to support very long-horizon memory. Some works have explored latent memory architectures~\citep{li2025cronusvla, shi2025memoryvla, fang2025sam2act, chung2025rethinking}, but only evaluated on short-horizon memory tasks. Others have proposed various heuristics to compress memory information, e.g., by relying on purely proprioceptive memory~\citep{zhang2025ta}, 2D point tracks~\citep{zheng2024tracevla, chen2025history}, by only retaining keyframes from prior timesteps~\citep{sridhar2025memer, wei2025cyclemanip, mark2026bpp}, or by representing memory in natural language~\citep{lin2025onetwovla}. %
A challenge for all these approaches is that it is hard to find a one-modality-fits-all solution for robot memory, and each individual representation will result in a compromise on capabilities: proprioception, point traces, or natural language alone, for example, lose precise spatial information about grasp angle and height that is necessary to correct a slipped grasp. Keyframes, on the other hand, need to aggressively sparsify the observation history to make inference computationally feasible in long-horizon tasks, but may lose the ability to estimate environment and robot dynamics, which often requires more densely sampled observations. 
In contrast to these prior works, we introduce a \emph{multi-modal} memory system that combines short-horizon, dense vision-based memory with long-horizon, language-based memory. This allows us to resolve partial observability, perform in-context adaptation, and solve long-horizon tasks, all without sacrificing efficiency. Finally, an orthogonal line of work proposes approaches for mitigating causal confusion~\citep{chi2025diffusion, zheng2024tracevla} in which a policy with memory erroneously learns to copy over prior actions, e.g., by introducing auxiliary objectives~\citep{torne2025learning}. While similar approaches \emph{could} be combined with our memory system, our experiments suggest that we can achieve high policy performance without such objectives, which may be attributed to our large-scale and diverse training data.

\noindent \textbf{Long-Context Models.} 
Outside of robotics, a large body of work has explored the training of long-context language and vision-language models \cite{liu2025comprehensive, tang2025video, ni2022expanding}. In particular, in the context of video processing, there is a rich literature on designing efficient encoders for long video inputs \cite{arnab2021vivit, bertasius2021space, li2024videomamba}. Our work leverages similar ideas to \cite{bertasius2021space, arnab2021vivit} for using sparse attention operations to efficiently process video inputs. Yet, in the context of robotics, latency constraints imposed by the real world require additional efficiency considerations: processing more than ten minutes of high-frequency, multi-frame video within a latency budget of a few hundred milliseconds is challenging, even on modern hardware. Thus, our work combines a short-horizon, video-based memory architecture with a significantly more compressed, language-based memory architecture for effective long-horizon context.

\section{Multi-Scale Embodied Memory for VLAs}
\label{sec:method}

In this section, we describe our proposed system for equipping VLAs with memory on multiple time scales.
For robot policies deciding on the next action to take, relevant context often spans multiple time scales.
The most recent timesteps help understand the dynamics of the robot and scene, resolve self-occlusions, and quickly adapt fine-grained manipulation strategies like re-grasps.
On the other hand, long-horizon memory is required to understand, for example, which steps of a recipe have already been completed, or which subtask has repeatedly failed.
In theory, both types of context could be provided by conditioning the policy on a dense sequence of all its previous observations. 
However, this becomes quickly infeasible for tasks spanning tens of minutes.
Our goal in designing \MethodName{} (\Acronym{}) is to enable both efficient short \emph{and} long-term memory.
For short-horizon memory, we use an efficient \textbf{video encoder} architecture that allows us to pass a multi-\emph{second} horizon of observations to the policy, enabling rapid in-context adaptation and robustness to self-occlusions.
For long-horizon memory, we design a \textbf{language-based memory} representation, and train the model to keep track of past events \emph{at a semantic level}, in natural language.
This way, we can equip our policies with memory spanning up to 15~minutes,
without sacrificing runtime latency constraints. %

We first give an overview of the \Acronym{} memory system (\cref{sec:system_overview}), then provide a detailed description of both the language-based memory (\cref{sec:language_memory}) and video encoder architecture (\cref{sec:video_encoder}), and finally detail the instantiation of our memory system in the \pizerosix{}-\Acronym{} VLA (\cref{sec:memory_vla}).

\subsection{Multi-Scale Embodied Memory (\Acronym{})}
\label{sec:system_overview}

\begin{figure*}
    \centering
    \includegraphics[width=\linewidth]{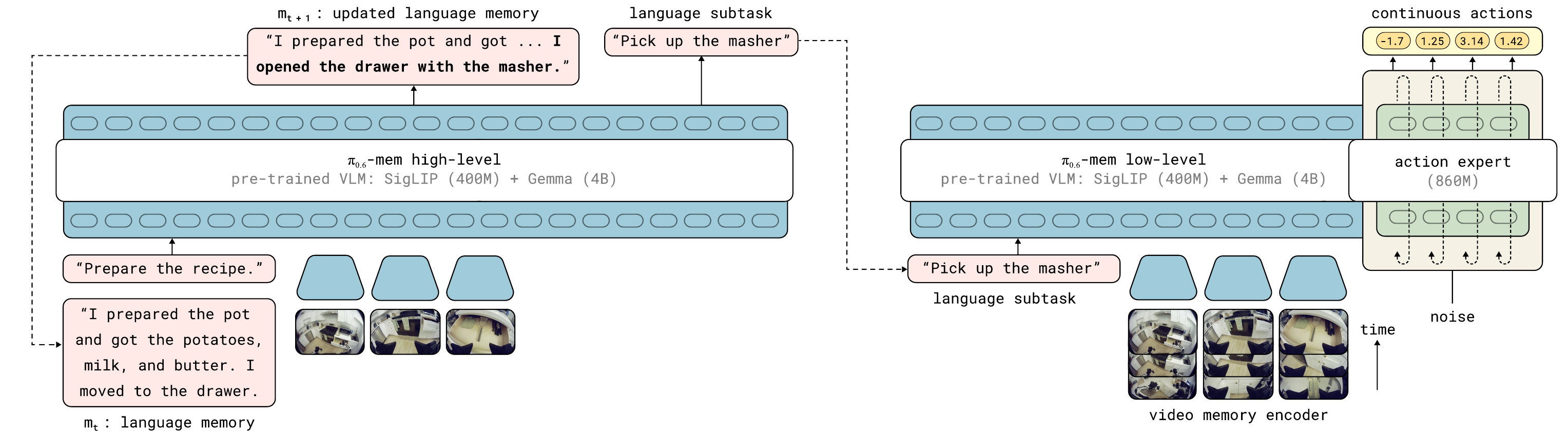}
    \caption{\small{The \Acronym{} memory system equips VLAs like \pizerosix{} with long-horizon memory via two key components: (1)~a high-level policy is trained to keep track of \textbf{long-horizon} semantic events by updating a \textbf{language memory} $m_t$ (left, \cref{sec:language_memory}), (2)~the low-level policy uses a \textbf{short-horizon} observation-based memory that is efficiently encoded via a \textbf{video encoder} (right, \cref{sec:video_encoder}).}}
    \label{fig:system_overview}
\end{figure*}

\cref{fig:system_overview} shows an overview of our \Acronym{} system.
Our goal is to train a policy $\pi(a_{t:t+H} | o_{t-T:t}, g)$ that predicts a chunk of continuous robot actions $a_{t:t+H}$ conditioned on the task goal $g$ in natural language, as well as a \emph{sequence} of dense observations (e.g. images, proprioceptive state) provided as input.
In comparison, most previous VLAs only condition on a single observation $o_t$, while we want the policy to process many ($T$) observations.

As mentioned above, scaling the number of past observations such that they span multiple minutes, and can thus serve as a form of long-range memory is infeasible.
Therefore, we factorize the problem of action prediction as follows:
\begin{align*}
 	\pi(a_{t:t+H},&\,l_{t+1}, m_{t+1} | o_{t-T:t}, m_t, g) \notag \\
 	\approx~&\pi_\text{LL}(a_{t:t+H} | o_{t-K:t}, l_{t+1}, g) \\
    &\pi_\text{HL}(l_{t+1}, m_{t+1} | o_{t}, m_{t}, g),
\end{align*}
where we separated the probability of actions into a low-level policy $\pi_\text{LL}$ and high-level policy $\pi_\text{HL}$.
The low-level policy models sequences of actions conditioned on the task goal $g$, a shorter sequence of observations ($K \ll T$) and a subtask instruction $l_{t+1}$.
The subtask instruction, in turn, is generated by the high-level policy, which is not only conditioned on the task goal, but also a summarization $m_t$ of previous semantic events in natural language.
We call this summarization the \emph{language memory} in the following.
It allows us to significantly reduce the number $K\ll T$ of dense observations that are input to the model, without sacrificing the ability to capture memory on the order of many minutes.
While previous work has employed a similar split into high- and low-level policies with subtask instructions as the interface, the key novelty here is that $\pi_\text{HL}$ additionally predicts the updated language memory $m_{t+1}$ based on its own previous prediction $m_t$.

\subsection{Language Memory for Long-Term Memory}
\label{sec:language_memory}
The language memory $m_t$ is a summary of previous \emph{semantic} events that happened while the policy is solving a task.
The main idea is to train the high-level policy $\pi_\text{HL}(l_{t+1}, m_{t+1} | o_{t-K:t}, m_{t}, g)$ such that it predicts the relevant summary $m_{t+1}$ from its previous summary $m_t$ and current observations and task.
This way, the model explicitly decides when and how to update its memory representation. For example, for a kitchen cleaning robot that already placed a plate in the cabinet and just finished picking up the next bowl, a memory update may look like the following:

\begin{center}
$m_t$: \small\texttt{I placed a plate in the cabinet and moved to the counter.}

\vspace{5pt}
$\Big\Downarrow$
\vspace{5pt}

$m_{t+1}$: \small\texttt{I placed a plate in the cabinet, moved to the counter, and picked up a bowl.}\hspace{-3pt}
\end{center}

In order to train $\pi_\text{HL}$, we need to obtain appropriate training data, including summarization actions we want the high-level policy to take. For this, we develop a pipeline to generate training data for the transition between $m_t$ and $m_{t+1}$ as follows.
Given a robot episode with subtask language annotations $l_{0:T}$, we pass the subtask instructions together with an indicator whether the subtask execution failed or succeeded to an off-the-shelf pre-trained language model (LLM) and ask it to summarize all information from the previous subtasks that is still relevant for future task execution. We collect the corresponding output and use it to label the sequence for training the high-level policy.

Importantly, we instruct the LLM to \emph{remove} or \emph{compress} information in the language memory whenever appropriate. For example, instead of remembering the precise attributes of all objects that were manipulated (``\texttt{I put a light green bowl, a dark blue bowl and a bright yellow bowl into the top right cabinet}''), it is often sufficient to just remember where the bowls were placed (``\texttt{I placed three bowls in the top right cabinet}''). Compressing the language memory helps keep it succinct (and thus inference fast) and reduces the potential for train-inference distribution shift, since fewer bits of information are carried over between timesteps. We find that current generation LLMs are effective at this task when prompted to keep only the minimal set of relevant information.

\subsection{Video Encoder for Dense Short-Term Visual Memory}
\label{sec:video_encoder}

\begin{figure}
    \centering
    \includegraphics{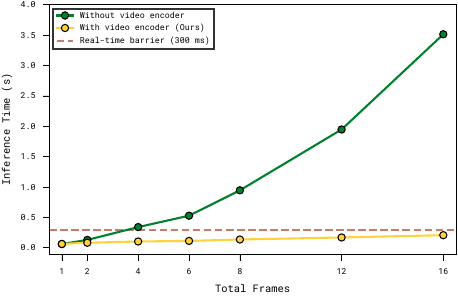}
    \caption{\small{Naively passing a sequence of observations into the backbone of a VLA rapidly increases inference latency. Our efficient video encoder architecture allows us to use many observation frames while remaining under critical real-time inference thresholds \cite{black2025real,black2025training}. Timings measured for the \pizerosix{} VLA, with four input camera streams on one NVIDIA H100 GPU.}}
    \label{fig:inference_time}
\end{figure}

The language memory mechanism, as described in the last section, can capture long-term \emph{semantic} concepts.
In order to enable our policies to reason about fine-grained details, dynamics, and resolve self-occlusion, access to a dense sequence of previous observations $o_{t-K:t}$ is required as input to the policy.
In most VLAs, encoding the observations, especially for images, accounts for the largest fraction of compute spent during training and significantly impacts inference speed.
Therefore, as we show in \cref{fig:inference_time}, simply encoding a sequence of observations one-by-one and passing them into the VLA backbone quickly becomes infeasible as the context grows beyond a few timesteps. The VLA will rapidly incur inference latencies beyond what prior work has deemed acceptable to prevent significant degradation in on-robot performance for dexterous manipulation tasks~\citep{black2025real, black2025training}.

\begin{figure}
    \centering
    \includegraphics[width=\linewidth]{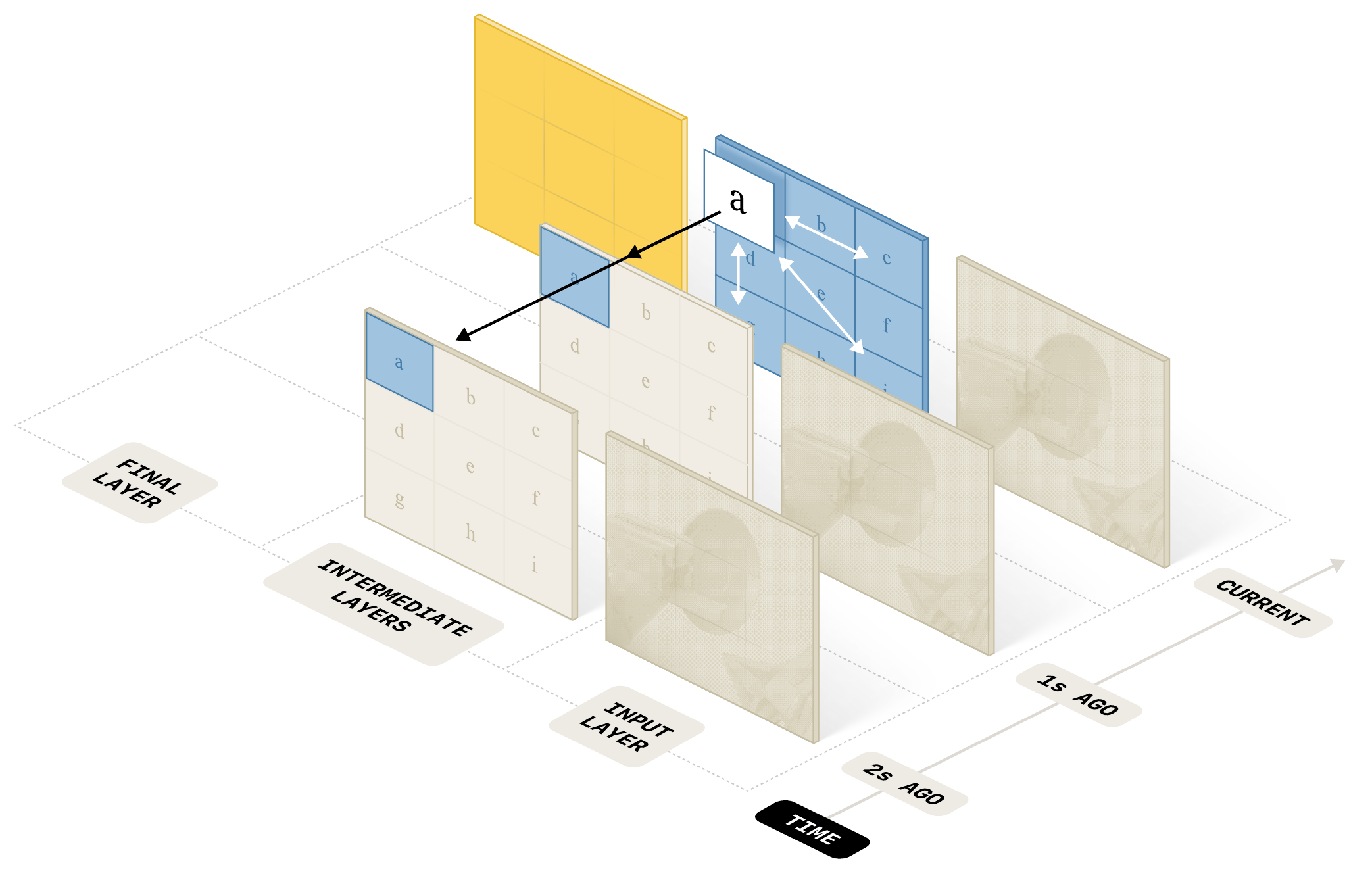}
    \caption{\small{We propose to use an efficient video encoder architecture for compressing short-horizon, image-based memory. Our architecture expands standard ViTs for encoding video inputs by interleaving layers that apply bidirectional spatial attention \emph{within} each observation (white arrows) with layers that additionally apply causal-temporal attention operations \emph{across} observations (black arrows). We drop observation tokens for past timesteps in upper layers of the ViT to compress the inputs and reduce the number of tokens passed to the VLA backbone.}}
    \label{fig:video_encoder}
\end{figure}

To address this, we propose to use a video encoder that can compress frames in the time dimension before passing the encoding into the VLM backbone.
Our video encoder extends Vision Transformers (ViTs)~\citep{dosovitskiy2020image}, typically used as the vision encoder in most VLAs, to encode video inputs.
Importantly, our encoder leaves the single-image performance of the VLM we start with invariant, and thus can be used to endow any pre-trained VLM with highly effective (yet computationally cheap) visual memory capabilities.

ViTs process images in patches and repeatedly perform bidirectional attention across the patch embeddings (across multiple layers) to derive a final set of output embeddings.
We keep the same general structure and our video encoder first patchifies all images in the input video separately. Inspired by existing work on space-time separable attention for video understanding (see e.g. \citet{bertasius2021space} for an overview) we then modify the attention mechanism every 4-th layer of the ViT to incorporate both spatial (as is standard in ViTs) but also temporal context. To avoid prohibitively expensive joint attention operations over the very large number of total patches in time and space, our architecture factorizes attention into separate \emph{spatial} and \emph{temporal} attention operations.
Every 4-th layer additively adds attention over the time dimension by performing attention over the timesteps' representations for \emph{the same} image patch using a causal attention mask (``temporal'') -- see \cref{fig:video_encoder} for a visual depiction and Appendix \ref{appendix:st_attention} for a mathematical description. 
This reduces the computational complexity of the corresponding attention in each layer from $\mathcal{O}(n^2K^2)$ (for naive attention over both time and space) to $\mathcal{O}(Kn^2 + n K^2)$.
Finally, to reduce the number of patches processed by the subsequent VLA transformer backbone, we only pass the representation computed for the current timestep onwards (dropping representations for all patches from past timesteps).
Thus, our video encoder \emph{matches} the number of tokens typically passed into the VLA backbone in single-timestep VLAs without memory; and we effectively force the video-encoder to incorporate temporal information into the representation produced for the current observation (via the modified attention mechanism).

A key property of our video encoder is that it \emph{does not} introduce new learnable parameters compared to standard, single-image ViTs. Video encoding capabilities are added by modifying the \emph{attention pattern} of the ViT and adding a fixed sinusoidal temporal position encoding. Thus, we can initialize the weights of our video encoder from the pre-trained ViT weights of any standard vision-language-models, just like in no-memory VLAs.
To maximize feature transfer, we ensure that for $K = 1$, i.e., single-image input, our encoder's initialization \emph{exactly} matches that of the VLM, which we achieve with a sinusoidal temporal position embedding that has value $0$ for $t = 0$.

In summary, our video encoder architecture allows us to effectively scale observation-based memory to tens of seconds without incurring prohibitive computational overhead during training or inference (\cref{fig:inference_time}), while at the same time permitting initialization from pre-trained vision-language model weights.

\subsection{Integrating \Acronym{} into the \pizerosix{} VLA}
\label{sec:memory_vla}

For our experimental evaluation, we equip the \pizerosix{} VLA~\cite{intelligence2025pi} with \Acronym{} memory by adapting its architecture to support the video encoder from \cref{sec:video_encoder}.
Like \pizerosix{}, the \pizerosix{} VLA with \Acronym{} is initialized from a pre-trained Gemma3-4B VLM~\citep{kamath2025gemma}.
Following~\citet{driess2025knowledge}, the model is trained with both discrete FAST action token prediction~\citep{pertsch2025fast} and a flow-matching action expert~\citep{blackpi0} with 860M parameters.
Gradients don't flow from the action expert into the VLM backbone~\citep{driess2025knowledge}. The model is trained with an input resolution of 448x448~px per camera stream and up to four camera streams (depending on the robot embodiment).

In addition to past camera frames, the observation memory of the \pizerosix{} model with \Acronym{} memory also contains past proprioceptive states, e.g., joint angles. The \pizerosix{} VLA represents the robot's state in text form. However, in the case of a \emph{sequence} of past states, this would quickly lead to a very large number of state text tokens. 
To prevent this, we instead use a continuous state embedding by projecting each proprioceptive state into the backbone embedding space with a linear projection.
This way, we only produce $K$ proprioceptive state tokens for an observation memory of length $K$.

We pre-train \pizerosix{}-\Acronym{} on a diverse mixture of data that includes teleoperated robot demonstrations, policy rollout data, and human corrections similarly to \citet{intelligence2025pi}, vision-language tasks, and \emph{video}-language tasks, such as video captioning.
During pre-training, we train the model with a sequence of six observations (five past and the current observation), spaced with a stride of one second.
During post-training, we find that we can flexibly expand this horizon and accommodate much longer observation-based memory (up to 18~frames and 54~seconds of observation-based memory in our experiments), similarly to what has been observed in language model and vision-language model training~\citep{chen2023extending}. 
We run all on-robot experiments using either inference-time real-time chunking~(RTC, \citep{black2025real}) or training-time RTC~\citep{black2025training} for asynchronous real-time inference.

\section{Experimental Evaluation}
\label{sec:experiments}

Our experimental evaluation aims to answer the following questions: (1)~Does \Acronym{} enable VLAs to perform tasks that require long-term memory spanning up to 15~minutes? (2)~Does \Acronym{} unlock new capabilities in VLA models, like in-context adaptation of manipulation strategies, based on short-term memory of recent failures? (3)~How does the performance of \Acronym{} compare to prior approaches for adding memory to VLA models?

\subsection{\Acronym{} Solves Tasks Requiring Long-Horizon Memory}
\label{sec:challenge_tasks}

\begin{figure*}
    \centering
    \includegraphics[width=\textwidth]{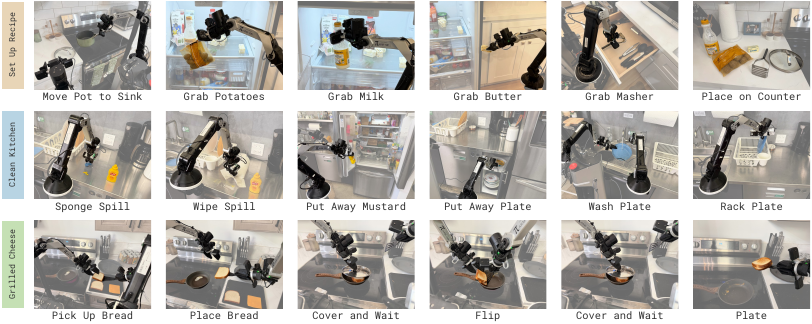}
    \caption{\small{We test \Acronym{} policies across multiple challenging, long-horizon dexterous manipulation tasks that require retaining memory for up to fifteen minutes, including setting up a recipe, cleaning up a kitchen (\cref{sec:challenge_tasks}), and making a grilled cheese sandwich (\cref{sec:analysis}).}}%
    \label{fig:challenge_tasks}
\end{figure*}

We evaluate our method's potential for enabling long-horizon tasks using two challenging scenarios that require retaining memory for up to fifteen minutes (see \cref{fig:challenge_tasks}, rows 1 \& 2): (1)~\textbf{Recipe setup}: The robot is given a detailed prompt specifying the ingredients and cookware required to cook a recipe and their location, and is asked to fetch all of them from various cabinets, drawers, or, e.g., the fridge, and place them at a specific location, e.g., on the stove or a specific countertop. Memory is required to keep track of all items that have already been assembled and remember to close drawers, cabinets, or the fridge once the task is complete. We train on 42~recipes across diverse kitchen scenes, and evaluate on five of the recipes in unseen kitchens and with unseen objects.
(2)~\textbf{Clean up kitchen}: The robot needs to clean up a messy kitchen environment, including stowing objects in the fridge, wiping countertops, washing dishes with soap and running water, and placing dishes in the drying rack. Memory is required for remembering steps in the cleanup process (e.g., whether soap has been added to the dishes before washing, whether both front and back of the dish have been washed), remembering which surfaces have already been cleaned, and remembering to close cabinets after objects have been stowed. This task is not only challenging from a memory perspective, but also requires policies to perform multiple dexterous manipulations, from wiping surfaces to carefully handling water and soap. For more details on training data, evaluation conditions, and scoring rubrics for progress scores, see \cref{sec:task_details}. All evaluations use \nchallengerollouts{}~rollouts per policy and task or recipe, and we report mean $\pm$ standard error in all graphs.

\begin{figure}
    \centering
    \includegraphics[width=\linewidth]{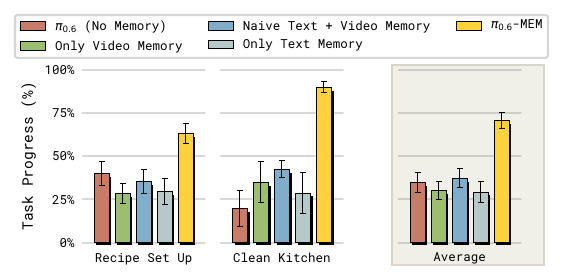}
    \caption{\small{Performance of policies on challenging, long-horizon manipulation tasks. Without memory, even state-of-the-art generalist policies like \pizerosix{} struggle to perform such tasks. We ablate the memory components and show these tasks are solvable by combining short-horizon, observation-based memory, with long-horizon, language-based memory. Naive memory of past language instructions, without compression, struggles with training-inference distribution shifts.}}
    \label{fig:challenge_tasks_results}
\end{figure}

\cref{fig:challenge_tasks_results} shows the results: without memory, even state-of-the-art models like \pizerosix{} struggle to perform these challenging long-horizon tasks. While prior works have shown policies that perform manipulation over extended periods of time~\citep{intelligence2025pi}, 
solving real-world tasks with flexible subtask sequences and frequent partial observability fundamentally requires memory. %
The results show that \Acronym{} is effective at providing the required context to the policy across both short and long-horizon time intervals, and increases policy success rate significantly.

To better understand what contributes to \pizerosix{}-\Acronym{}'s strong performance, we perform a detailed ablation study of the different components of our approach. We report the results in \cref{fig:challenge_tasks_results}. We compare to versions of our policy that \emph{remove} short-horizon video memory or long-horizon language memory, and show that both are essential to solving the challenge tasks. Without video memory, the robot may struggle to understand how long it has been washing a plate or wiping a surface, and get ``stuck'' indefinitely. Video memory also helps robustify manipulation tasks via in-context adaptation (see \cref{sec:in_context_exp}). Long-horizon language memory, on the other hand, is essential for remembering semantic events in the more distant past. It allows the policy to keep track of steps in the recipe, or remember to close doors that it has opened. We also evaluate a version of our policy that removes the \emph{compression} applied to the language-based memory (see \cref{sec:language_memory}): instead of training the model to compress and discard information that is no longer needed, we simply concatenate all previous subtask instructions up to a maximum length in the input of the high-level policy. We find that this type of ``naive'' language memory works significantly worse than our model-predicted summaries. The core challenge with ``naive'' language memory is a large train-inference distribution shift: during training, most episodes utter any given subtask instruction only once (e.g., ``\texttt{pick up bowl}'' $\rightarrow$ ``\texttt{place bowl in cabinet}'') since they are typically near-optimal human demonstrations. Yet, during inference time policies may repeatedly fail on a given subtask, causing the high-level policy to repeatedly produce the same subtask before finally succeeding and moving on (``\texttt{pick up bowl}'' $\rightarrow$ ``\texttt{pick up bowl}'' $\rightarrow$ ``\texttt{pick up bowl}'' $\rightarrow$ ``\texttt{place bowl in cabinet}''), leading to a distribution shift that can degrade overall policy performance. In contrast, \Acronym{}'s language memory mechanism would simply not update the memory representation until the bowl was successfully picked up. This \emph{compression} of context (e.g., discarding failed attempts) thus reduces distribution shift and improves overall performance. 

\subsection{In-Context Adaptation of Manipulation Strategies}
\label{sec:in_context_exp}

\begin{figure}
    \centering
    \includegraphics[width=\linewidth]{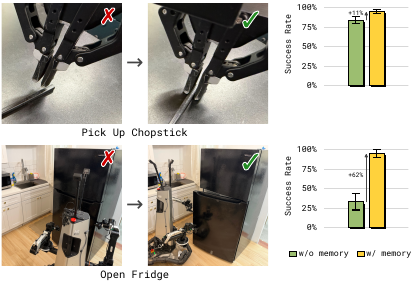}
    \caption{\small{VLAs with memory can perform in-context adaptation of manipulation strategies, like adjusting grasp height or door opening direction. Without memory, policies get stuck with a suboptimal strategy.}}
    \label{fig:in_context_adapt}
\end{figure}

In the previous section, we showed that \Acronym{} can enable VLAs to solve very long-horizon tasks. In this section, we investigate whether equipping VLAs with memory can unlock improved performance, even on shorter-horizon tasks that at first glance may not require memory. Intuitively, while memory across tens of minutes is useful for keeping track of overall task progress, short-horizon memory can enable policies to \emph{adapt} their behavior \emph{in-context} and intelligently react to mistakes: instead of failing in the same way over and over, policies can use context of previous failed attempts to, for example, modify the way in which they are trying to pick up an object, or how they open a door. 

To test whether \Acronym{} unlocks such in-context adaptation, we set up two tasks on which current state-of-the-art VLAs like \pizerosix{} struggle (\cref{fig:in_context_adapt}): picking up flat objects like chopsticks with an out-of-distribution table height, which leads to frequent mis-grasps, and opening fridges where the direction the door opens is unclear, resulting in repeated failed opening attempts.

To teach policies the in-context adaptation strategy, we follow~\citep{intelligence2025pi} and collect targeted human feedback: after the policy fails, a human intervenes and provides a demonstration of the corrected manipulation strategy, like adjusting grasp height for picking up the chopstick. For the fridge-opening task, we collect exploration rollouts in which the demonstrator initially does not know the opening mechanism, so the data naturally contains both the failed attempt and the subsequent corrective demonstration. We then simply finetune the \pizerosix{}-\Acronym{} policy with this correction data, keeping the failed attempt that preceded the correction in the short-term memory of the model during training. As a result, the model learns to adapt its manipulation strategy when it sees a mistake in its short-term memory. We compare to finetuning the memoryless \pizerosix{} policy on the same intervention data.

The results in \cref{fig:in_context_adapt} show that the \Acronym{}-VLA with memory is much more effective at leveraging the corrections, and learns to adapt its manipulation strategy on the fly. The policy without memory has no way to remember which strategy was attempted before, and therefore cannot intelligently change the strategy after a mistake.
In contrast, policies with memory can use the context to understand which strategy has already been tried and failed, and adjust accordingly.

\subsection{Analysis Experiments}
\label{sec:analysis}

\begin{figure*}[t]
    \centering
    \includegraphics[width=\textwidth]{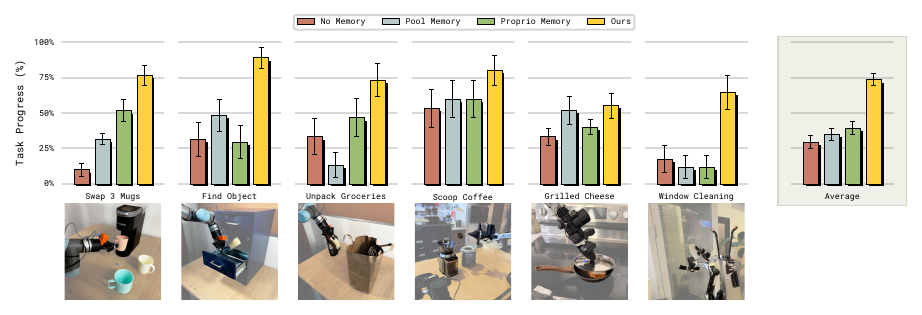}
    \caption{\small{Comparison of approaches for equipping VLAs with memory across core memory capabilities (handling partial observability, counting, visual memory). VLAs without memory struggle to perform these tasks. Only \Acronym{}'s memory approach performs well across all core capabilities.}}
    \label{fig:analysis_results}
\end{figure*}

\begin{figure*}[t]
    \centering
        \includegraphics[width=\textwidth]{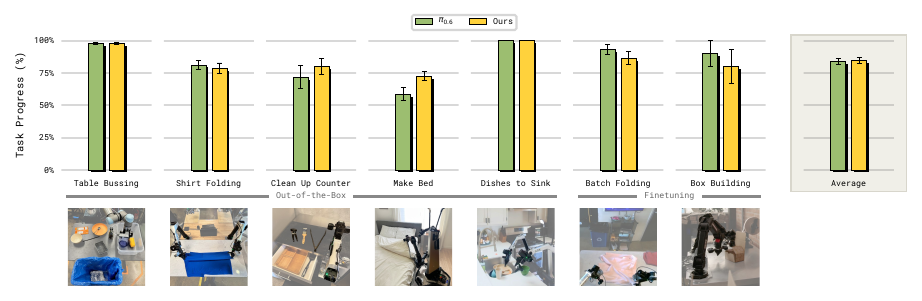}
    \caption{\small{In addition to effectively using its memory, \Acronym{} also matches the performance of state-of-the-art non-memory VLAs across challenging manipulation tasks that do not require memory.}}
    \label{fig:dexterous_results}
\end{figure*}

\textbf{Tasks.} To compare our method to existing approaches that equip VLAs with memory across a wide range of capabilities, we develop a suite of challenging manipulation tasks that measure the ability of policies to use memory efficiently, and perform dexterous manipulations (see \cref{fig:analysis_results,fig:dexterous_results} for task visualizations). %
The tasks span a diverse set of scenarios involving single-arm, dual-arm, and mobile robots. They test for core memory capabilities, like handling \textbf{partial observability} (e.g., remembering which drawer an object was placed in by a human; unpacking a grocery bag and remembering whether objects are left to unpack; placing and removing multiple mugs under a coffee machine), \textbf{counting} (e.g., counting scoops of coffee to add to a coffee grinder), timing (e.g., remembering how long a grilled cheese sandwich has been cooking, see \cref{fig:challenge_tasks}), and \textbf{spatial memory} (e.g., remembering which parts of a window have already been wiped). They also require policies to perform \textbf{precise manipulations} (e.g., folding a stack of laundry or assembling a cardboard box). They are thus well-suited to test the limits of existing memory approaches and also compare to the performance of state-of-the-art VLAs without memory. Following prior work~\citep{blackpi0}, we collect small post-training datasets for the most challenging manipulation tasks and for the memory tasks, and evaluate policies out-of-the-box for all remaining tasks (\cref{fig:dexterous_results}).
We provide detailed descriptions of all task setups in \cref{sec:task_details}.

\textbf{Comparisons.} A number of prior works leverage observation-based (shorter-horizon) memory, and address its computational challenges through aggressive compression of observations from past timesteps. We compare our approach to two representative prior works to understand the tradeoffs of different memory representations: (1)~\textbf{Pool Memory} akin to \citet{jang2025contextvla}, we
compress all past observations into a single ``memory token'' using average pooling. We implement it by encoding each of the past observations separately with the pre-trained single-frame ViT, and then applying an average pooling operation on the frame encodings before passing them into the VLA backbone. The current timestep's observation is separately encoded and passed into the VLA without pooling; (2)~\textbf{Proprio Memory} conditions on a history of low-dimensional robot states only, akin to~\citep{zhang2025ta}, to avoid the computational cost of high-dimensional image memory. All proprioceptive states from past timesteps are separately passed through a linear projection, and then fed into the VLA backbone. To ensure fair comparison, we re-implement all approaches on top of the \pizerosix{} VLA backbone and use a version of our model \emph{without} language-based long-horizon memory. To test the performance of our policy on the most dexterous tasks, we also compare to \textbf{\pizerosix{}} \emph{without} memory, a state-of-the-art VLA that has shown strong results on complex manipulation tasks~\citep{pi06_model_card}. Finally, to better understand the effect of \emph{pre-training} on memory efficacy, we compare to \textbf{\Acronym{}-Posttrain-Only}, an ablation of our method that introduces the video encoder only during \emph{post-training} on the target task, starting from the pre-trained \pizerosix{} checkpoint, akin to \citet{li2025cronusvla}.

We report results in \cref{fig:analysis_results}. We first compare the different memory approaches on tasks that test core memory capabilities (\cref{fig:analysis_results}). As expected, the \pizerosix{} VLA \emph{without} memory struggles on these tasks and often has to resort to random chance, e.g., when picking which of four drawers to open to find the object hidden inside (25\% success) or whether or not to add another scoop of coffee (50\% success). In contrast, existing memory solutions improve performance, particularly on simpler memory tasks that only require a few bits of memory, e.g., remembering how many scoops of coffee have already been added to the hopper. We find that Pool-Memory's aggressive observation compression via average-pooling can lead it to struggle particularly on tasks that require longer-term memory, like remembering the past positions of multiple coffee mugs or recalling how many objects are left to unpack in a grocery bag. Proprio-Memory on the other hand is only effective in tasks where the robot needs to remember \emph{its own} state, but struggles in scenarios where remembering the \emph{environment} state is necessary: e.g., recalling which drawer contains a particular object.

\begin{figure}
    \centering
    \includegraphics[width=\linewidth]{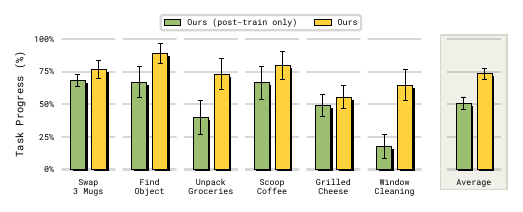}
    \caption{\small{Pre-training \Acronym{} on a diverse dataset of robot and non-robot video data makes it more effective at using its memory to solve diverse manipulation tasks. Introducing memory during post-training only results in substantially lower performance.}}
    \label{fig:posttrain_ablation}
\end{figure}

In comparison, the \Acronym{} VLA is the only model that achieves strong performance across \emph{all} tested memory capabilities. It can reliably handle partial observability challenges like remembering the location of objects, and also leverage these capabilities in dexterous tasks like unpacking a bag of groceries. Notably, pre-training the observation-based memory on a diverse data mix of robot and non-robot data significantly improves the ability of the model to leverage its memory -- even when the memory horizon is significantly expanded during post-training (e.g., from 5~seconds in pre-training to up to 1~minute in post-training). The version of our model that only introduces memory during post-training is noticeably worse at leveraging the information from past timesteps (see \cref{fig:posttrain_ablation}). We emphasize that this model is still initialized from a base \pizerosix{} checkpoint pre-trained on the same diverse robot data, but did not develop memory capabilities during this pre-training. Finally, the head-to-head comparison of Ours (post-train only) vs. Pool-Memory demonstrates that even without pre-training, our video encoder design is a more effective approach for extracting information from prior timesteps than separately encoding each timestep and average pooling. By interleaving temporal attention operations throughout the layers of the ViT, our video encoder can more effectively extract and compress the relevant information.

In addition to testing core memory capabilities, we also test the performance of \Acronym{} across a range of challenging dexterous manipulation tasks (\cref{fig:dexterous_results}). We find that the \Acronym{} VLA not only makes effective use of memory in tasks that require it, but it can \emph{also} match the state-of-the-art performance of the \pizerosix{} VLA across challenging dexterous manipulation tasks. This is notable, since numerous previous works have reported degradations in performance from adding memory to policies, e.g., due to causal confusion~\citep{chi2025diffusion, zheng2024tracevla}. We attribute \Acronym{}'s strong performance in large parts to our diverse pre-training data mixture, which contains episodes of varying optimality, speed, and control frequency, in addition to diverse internet videos. Together, this diversity can prevent spurious correlations in the video encoder that may arise when training on smaller, more uniform robot datasets, and lead to the brittleness and performance degradation of policies with memory reported in prior work. %

\section{Conclusion}
\label{sec:conclusion}

We presented Multi-Scale Embodied Memory (\Acronym{}), an approach for equipping VLAs with long-horizon memory. \Acronym{} enables VLAs to perform long-horizon tasks that require tens of minutes of memory, while obeying real-world latency constraints. \Acronym{} also enables in-context adaptation of manipulation strategies and, when combined with the \pizerosix{} VLA, achieves state-of-the-art performance across a wide range of manipulation tasks. To achieve this, \Acronym{} introduces a mixed-modal memory architecture that combines short-horizon, video-based memory, with a long-horizon, language-based memory mechanism. 
We believe that \Acronym{} is only the first step towards building robot policies that can effectively manage very long-horizon memory. Future work can explore how we can scale memory to last beyond the horizon of a single episode, to span weeks, months, or years of deployment, and allow us to build robots that learn continually at deployment time.

\section*{Acknowledgements}
Like any robotics effort of this scale, our work involved a large team of people that helped with data collection, evaluation, robot operations, and maintenance. We thank the Pi robot operators and annotators for their help with data collection, evaluation, and annotation. We thank the operations team for help with logistics and scheduling. And we thank the hardware team for keeping all our robots running. We also thank Lili Yu for helpful discussions about the video encoder architecture. We thank Claudio Guglieri for help with visualizations for the blog post, and Donald Jewkes for help with filming and editing of the video.

\bibliographystyle{plainnat}
\bibliography{references}

\newpage
\onecolumn
\appendix

\subsection{Contributions}
The project was started by HW and DD. DD, HW, and JTS designed the video encoder and short-horizon memory system. KP designed the long-horizon memory system. MT, KP, HW, KV, SN, and ME designed the evaluation suite and performed experimental evaluations, including model post-training. DD, KP, AR, and QV developed the pre-training recipe and data mixture, and ran large-scale model training. HW, KD, KS, and MT developed the infrastructure for long-context VLA training. JT optimized inference of the \Acronym{} VLA. SL and CF provided advice throughout the project. MT, KP, DD, BI, KV, SN, SL, and CF worked on writing, illustrations, and the video. 

\subsection{Task Details}
\label{sec:task_details}

\subsubsection{Long-horizon Tasks (\cref{sec:challenge_tasks})}

\paragraph{Preparing the items for a recipe} The goal of the task is for the robot to take out all of the items for a recipe. The items are placed in various, randomized locations throughout the kitchen (fridge, cabinet, drawers, stove ...). The policy's prompt specifies which items should be assembled, where they should be placed (countertop, stove, sink), and their respective initial location in the kitchen. We list a few prompt examples below. The robot will receive 1 point per item retrieved and placed in the requested position. Most recipes consist of 6 to 7 points. We collected data on over 40 recipes with a large variety of objects, scenes, appliances, prompts. The evaluation is done on a variety of unseen scenes, with seen recipes. 
\begin{PromptBox}{Mashed potatoes}
Take out supplies for the recipe. Remove the lid from the pot on the stove and place the lid on the countertop to the left of the sink, and pick up the pot and place it in the sink. Get the potatoes, butter, milk from the fridge and place them on the countertop to the left of the sink. Get the masher from the top drawer to the left of the sink and place it on the countertop to the left of the stove.
\end{PromptBox}

\begin{PromptBox}{Fried rice}
Take out the supplies for the recipe. Get the bag of rice and the spam from the cabinet on the bottom right of the dishwasher and place them on the countertop to the left of the sink. Get the soy sauce from the fridge and place it on the same section of the countertop to the left of the sink. Get the pan from the bottom cabinet to the right of the dishwasher and put it on the stovetop. Get the spatula from the top drawer to the right of the dishwasher and place it on the pan on the stovetop.
\end{PromptBox}

\begin{PromptBox}{Pizza}
Take out the supplies for the recipe. Get the baking tray from the bottom cabinet on the left of the dishwasher and place it on the countertop to the left of the sink. Get the pizza dough, package of pepperoni, and bag of cheese from the fridge and place them on the countertop to the left of the sink. Get the dough roller from the top drawer to the right of the stove and place it on the countertop to the left of the sink.
\end{PromptBox}

\paragraph{Cleaning the kitchen} This task requires cleaning a kitchen by wiping the countertop with the sponge, drying the countertop with a paper towel, storing any food items left in the countertop inside the fridge, putting all the dishes from the dishrack into the cabinets, and washing all of the dishes from the sink. The scoring consists of +1 point per subtask done and on average episodes consist of 8 subtasks. Below we provide an example prompt.
\begin{PromptBox}{Clean kitchen}
Clean up the kitchen. Wipe the countertop with the sponge and dry it with the paper towel and throw the towels in the trash can under the sink. Put the mustard in the fridge. Put the dishes in the bottom cabinet on the left of the fridge. Wash the blue plate and black plate and place them in the dishrack. 
\end{PromptBox}

\subsubsection{In-context adaptation (\cref{sec:in_context_exp})}

\paragraph{Chopstick Pick Up} The policy is tasked with picking up a chopstick on a variable height table. During data collection, the chopstick is placed at random locations on the table with random table heights in the upper half of the table height range. We collect human interventions whenever the policy mis-grasps, and then train both memory and non-memory policies on this data and evaluate them with the chopstick randomly placed on the table at the lowest table height setting. The final policies are scored +1 for picking up the chopstick and +1 for placing it in the bin, and success is a score of 2.

\paragraph{Open Refrigerator} The goal of the task is to open a fridge in front of the robot. The fridge does not have obvious visual cues indicating which side the door hinge is on, leading policies to often attempt to open it in the wrong direction. We collect human interventions whenever the robot opens the fridge in the wrong direction. For evaluation, an episode is defined as successful if it takes $\leq 4$ grasps to open the door, to test for intentional strategy switching rather than repeated random sampling.

\vspace{0.3cm}
\subsubsection{Analysis Experiments (\cref{sec:analysis})}
\paragraph{Three-way swap mugs} The goal is to place three coffee mugs under a coffee machine sequentially. Two mugs start at random positions on the table and a third mug starts under the coffee machine. The scoring consists of obtaining +1 point for each mug that goes on the coffee machine without tipping and without repeating mugs. Only one mug can go on the coffee machine at a time, so the robot needs to place it back on a random location on the table after it was placed under the coffee machine to make space for the next mug.

\paragraph{Find object} The goal is to retrieve the hidden object in a cabinet with four drawers. In the beginning of the episode, a person places the object in a random drawer. The robot needs to remember which drawer the person placed the object into, then open the correct drawer and retrieve the item and place it on the table. One point will be received for successfully accomplishing the task without opening any incorrect drawer.

\paragraph{Unpack groceries} The goal is to retrieve all items from a grocery bag without missing any items inside. The inside of the bag is not observable from the third person camera of the robot, and the number of items in the bag is randomized across episodes. Thus, to complete the task, the robot will need to remember how many objects are left in the bag from its past wrist camera observations. Episodes are marked as a success if the policy retrieved all items from the bag and indicated completion by moving home once all items are retrieved.

\paragraph{Scoop coffee} The goal is to put exactly two scoops of coffee beans in a grinder. The robot needs to remove the lid from the hopper of the grinder, pick up a coffee scoop, put two scoops of coffee from an opened coffee bean package into the grinder, then put the lid back on. Episodes are marked as a success if exactly two scoops are added to the grinder and the lid is put back on.

\paragraph{Grilled cheese} The goal of the task is to prepare a grilled cheese sandwich. The robot has to assemble the grilled cheese sandwich in the pan, then wait for the grilled cheese to cook, flip it, wait for the other side to cook, and finally plate the grilled cheese. The policy receives +1 point for assembling the sandwich, cooking for the correct amount of time (between 30 seconds and 3 minutes per side), flipping, and plating.

\paragraph{Window cleaning} The goal is to wipe the window door of a phone booth. The task consists of spraying windex on the window, ripping off a paper towel from a roll, then wiping the whole window, and finally throwing the paper towel in the trash. The robot needs to remember which steps have been completed and which part of the window have already been wiped during the cleaning phase. Episodes are marked as a success if all steps of the task are completed. It is considered a failure if the policy fails to wipe parts of the window.

\paragraph{Table bussing} The goal is to sort 12~objects from a tabletop into (A)~a bussing bin (for utensils and tableware), and (B)~a trash can (for plastic containers, bottles, and wrapping paper). The policy receives a score of +1 for every correctly placed item.

\paragraph{Shirt folding} The robot has to fold a shirt on a tabletop. At the beginning of the episode the shirt is flattened on the table. The episode is scored as a success if the shirt is successfully folded into a rectangle without seams sticking out or excessive wrinkles.

\paragraph{Clean up counter} The goal is to clean items from a counter into a drawer. The robot has to open the drawer, place all items on the counter into the drawer, and close it. An episode is counted as a success if all subtasks are completed successfully.

\paragraph{Make bed} The goal is to tidy a bed, by straightening the blanket, and placing any pillows at the head of the bed. The episode is scored as a success if all pillows are placed at the head of the bed, and the blanket is tidied.

\paragraph{Kitchen cleanup} The goal is to clean the counter of a kitchen, by placing multiple plates, bowls, and a cutting board from the counter into a nearby sink. The episode is scored as a success if all plates are placed into the sink.

\paragraph{Batch folding} In contrast to the shirt folding task, here the clothes start in a hamper on one side of the table. The robot needs to take the clothing item out of the hamper, flatten it, fold it, and then stack it onto a stack of existing clothes on the table. The hamper may contain various t-shirts and shorts. Episodes are scored as successes if the clothing items are folded into rectangles without seams sticking out or excessive wrinkling.

\paragraph{Box building} Given a flattened cardboard cutout, the robot is tasked to fold the cardboard into a box. This task requires multiple precise, well-coordinated bimanual manipulations to assemble the box. Episodes are scored as successful if the box is fully assembled without damaging the box.

\subsection{Video encoder with Space-Time separable attention}
\label{appendix:st_attention}
We describe how we can adjust a layer in a given ViT image encoder to instantiate our video-encoding scheme.

Let $\mathbf{z}^{l-1}_{p, t}$ denote the input embeddings to layer $l$ for spatial patch $p$ and timestep t (where $t \in [-K, 0]$). We first modify the embedding by adding a sinusoidal position embedding based on $t$, denote the output of this step with 
$$
\hat{\mathbf{z}}^{l-1}_{p, t} = \mathbf{z}^{l-1}_{p, t} + e(t),
$$
where $e(t)$ denotes the sinusoidal position embedding and we set the boundary condition $e(0) = 0$.

We then re-use the ViT's standard query key and value projections which in our case are given as
\begin{equation}
    \begin{aligned}
        \mathbf{q}^{l, a}_{p, t} &= W^{l, a}_Q \text{LN}\big( \hat{\mathbf{z}}^{l-1}_{p, t} \big), \\
        \mathbf{k}^{l, a}_{p, t} &= W^{l, a}_K \text{LN}\big( \hat{\mathbf{z}}^{l-1}_{p, t} \big), \\
        \mathbf{v}^{l, a}_{p, t} &= W^{l, a}_V \text{LN}\big( \hat{\mathbf{z}}^{l-1}_{p, t} \big),
    \end{aligned}
\end{equation}
where $a$ indexes the attention head and LN denotes a layer-norm implementation (we use RMSNorm but this choice is dependent on the ViT we start from).

Next, we can define the general attention mechanism (ignoring normalization for brevity of notation) as the softmax over the query and key outer product:
\begin{equation}
    \mathbf{\alpha}^{l,a}_{p,t} (\mathcal{S} = \lbrace1, \dots N \rbrace, \mathcal{T} = \lbrace 1, \dots, T \rbrace) [\hat{\mathbf{z}}] = \text{SM}\Big( (\mathbf{q}^{l, a}_{p, t})^T \cdot \big[ \mathbf{k}^{l, a}_{0, 0} \lbrace \mathbf{k}^{l, a}_{p', t'} \rbrace_{p' \in \mathcal{S}, t' \in \mathcal{T}} \big]\Big),
\end{equation}
where $\text{SM}$ denotes the softmax operation and $\mathcal{S} = \lbrace1, \dots N \rbrace$ and $\mathcal{T} = \lbrace 1, \dots, T \rbrace$ denote the space and time indices we want to perform attention over respectively.
With this general definition we can define the space only attention mechanism as instantiating $\mathbf{\alpha}^{l,a}_{p,t} (\mathcal{S} = \lbrace 1, \dots N \rbrace, \mathcal{T} = \lbrace \rbrace)$ and the time only attention mechanism as instantiating $\mathbf{\alpha}^{l,a}_{p,t} (\mathcal{S} = \lbrace \rbrace, \mathcal{T} = \lbrace 1, \dots, T \rbrace)$.
Thus the attention mechanism used in our video encoder is precisely given as
\begin{equation}
    \mathbf{\alpha}^{l,a}_{p,t} (\mathcal{S} = \lbrace 1, \dots N \rbrace, \mathcal{T} = \lbrace \rbrace) [ \mathbf{\alpha}^{l,a}_{p,t} (\mathcal{S} = \lbrace1, \dots N \rbrace, \mathcal{T} = \lbrace 1, \dots, T \rbrace)[\hat{\mathbf{z}}]].
\end{equation}
And thereafter we follow the standard computation of a transformer layer, i.e. we compute the outputs of the transformer by first combining attention weights $\alpha$ with values $\mathbf{v}$ and passing the corresponding outputs to a MLP.

\end{document}